%% file: iros20gonzalez.tex
\documentclass[letterpaper, 10 pt, conference]{style/ieeeconf}

\IEEEoverridecommandlockouts

\usepackage{cite}
\usepackage{amsmath, amssymb, bm}
\usepackage{graphicx}
\graphicspath{{figs/}}	
\usepackage{xcolor}		
\usepackage{pdfpages}

\newcommand{\executeiffilenewer}[3]{%
	\ifnum\pdfstrcmp{\pdffilemoddate{#1}}%
	{\pdffilemoddate{#2}}>0%
	{\immediate\write18{#3}}\fi%
	}

\title{Line Walking and Balancing for Legged Robots with Point Feet}

\author{Carlos Gonzalez$^{1}$, Victor Barasuol$^{1}$, 
		Marco Frigerio$^{2}$, Roy Featherstone$^{3}$,\\ 
		Darwin G. Caldwell$^{3}$,
		Claudio Semini$^{1}$%
\thanks{$^{1}$Dynamic Legged Systems Lab, 
	Istituto Italiano di Tecnologia, Genoa, Italy.
	Email: $<$first name$>$.$<$last name$>$@iit.it}
\thanks{$^{2}$Department of Mechanical Engineering, KU Leuven, Belgium. Email: marco.frigerio@kuleuven.be}
\thanks{$^{3}$Department of Advanced Robotics, 
	Istituto Italiano di Tecnologia, Genoa, Italy.
	Email: $<$first name$>$.$<$last name$>$@iit.it}}%

\newcommand{\RF}[1]{\textcolor{red}{#1}}	%
\begin{document}

\null%
\includepdf[pages=-]{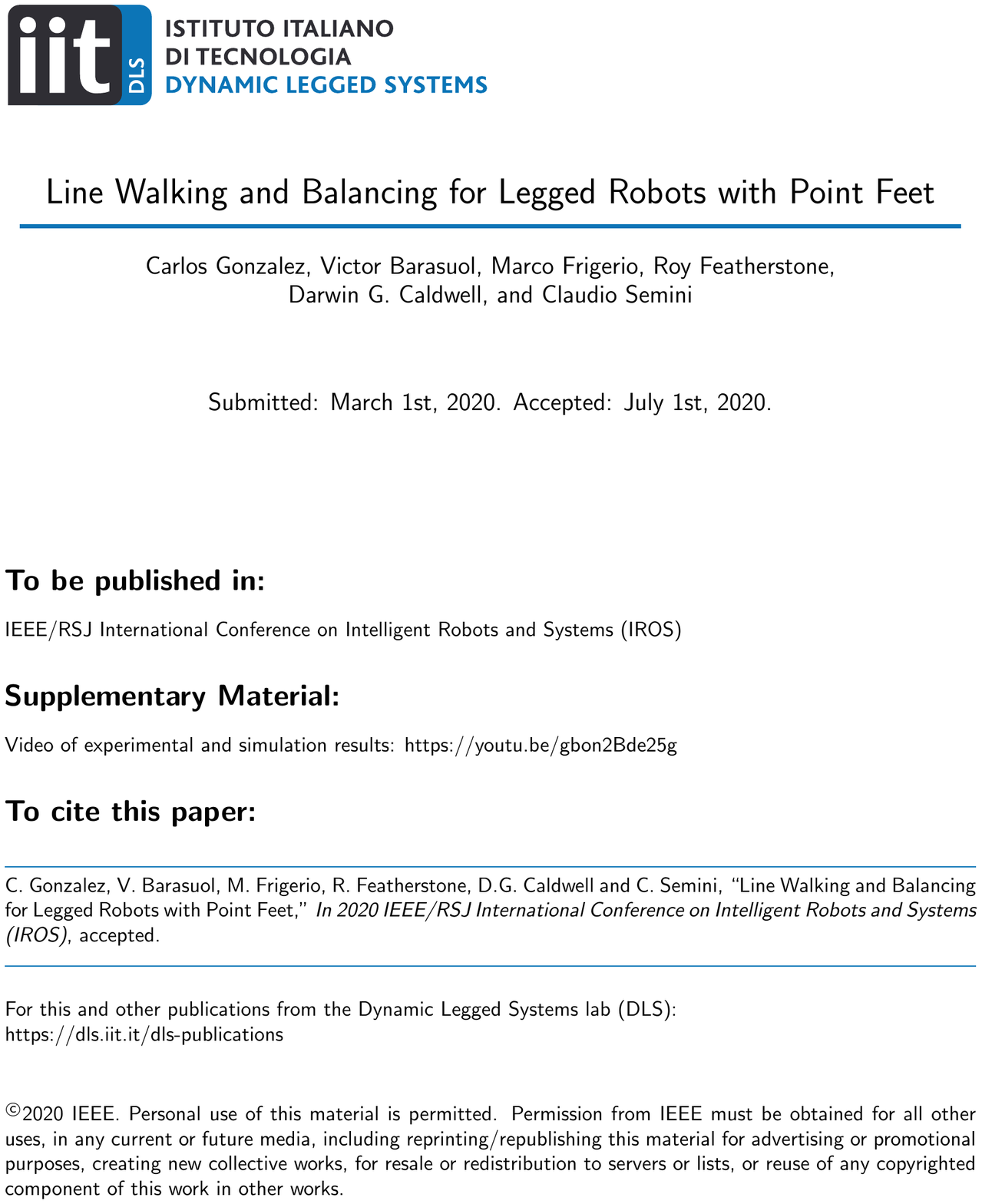}

\maketitle	
\thispagestyle{empty}
\pagestyle{empty}

\begin{abstract}
The ability of legged systems to traverse 
highly-constrained environments depends by and large 
on the performance of their motion and balance controllers. 
This paper presents a controller that
excels in a scenario that most
state-of-the-art balance controllers have not yet addressed: 
line walking, or walking on nearly null support regions.
Our approach uses a low-dimensional
virtual model (2-DoF) to generate balancing actions through
a previously derived four-term balance controller and 
transforms them to the robot through a derived
kinematic mapping.
The capabilities of this controller are tested in simulation, 
where we show the 90kg quadruped robot
HyQ crossing a bridge of only 6 cm width
(compared to its 4 cm diameter spherical foot), 
by balancing on two feet at any time while moving along a line.
Additional simulations are carried to test
the performance of the controller and the 
effect of external disturbances.
Lastly, we present our preliminary experimental results showing 
	HyQ balancing on two legs while being disturbed.
\end{abstract}


\section{Introduction}
\label{sec:introduction}
Legged robotic systems are becoming increasingly versatile 
and their capabilities are being evaluated in
fields such as inspection and disaster relief, among others. 
The usability of these systems, however, is
compromised when the robot encounters an environment 
with limited support contacts for the feet. 
In these scenarios, the balancing capabilities of the robot are crucial for the completion of the task.

\begin{figure}[b]
	\centering
	\includegraphics[width=0.8\linewidth]{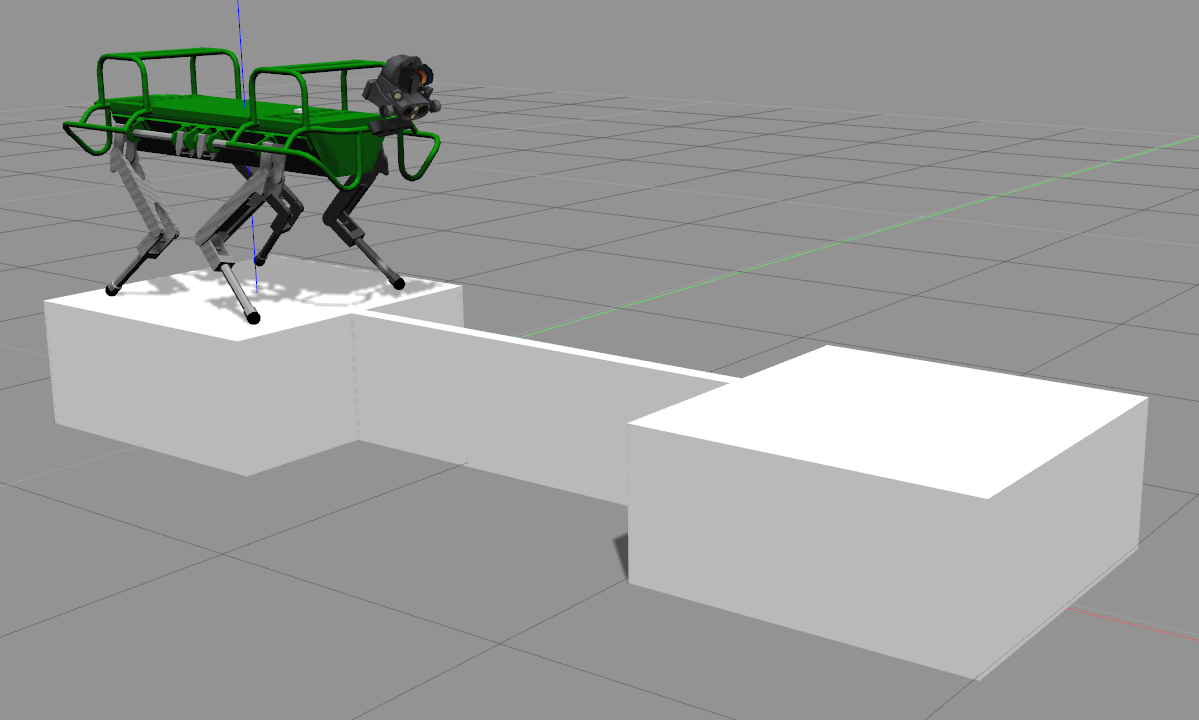}
	\caption{Sample scenario where a high-fidelity balance 
		controller is required. The length and width of the bridge
		are 1.5 m and 6 cm, respectively.}
	\label{fig:narrow_beam}
\end{figure}

%
Consider the scenario of crossing a narrow bridge, as shown in
Fig. \ref{fig:narrow_beam}. In this case, a high-fidelity 
balance controller is 
required. Most of today's balance controllers, however, would fail this task since line walking on a nearly null support polygon is required.
In this work, we consider the possibility
of achieving this task (crossing a narrow bridge) 
using a momentum-based balance controller applied
on a low-dimensional virtual model
that controls the physical process of balancing.


This work is based on the balance controller of
\cite{Featherstone2017}, which was originally for planar systems, 
but in this paper we will extend this approach to the control 
of a 12 degrees of freedom (DoF) quadruped, which will balance 
on two point feet. In addition, we will further extend the 
ideas previously published in \cite{Featherstone2017}
to allow motions along a contact line in 3-dimensions.
To the best of our knowledge, this is the first time that such 
a task has been achieved in simulation and that a heavy legged robot 
has balanced on a null support polygon experimentally 
without continuously taking steps.

The main contributions of this work are 
(a) 
the derivations of the kinematic mapping
relating the planar model and a quadruped, 
(b) 
the usage of the extended balance controller in conjunction
with an external motion generator to create balanced 
walking motions, 
(c)
simulation experiments showing the performance 
of the controller tracking an  
angular position reference
on a quadruped,
(d)
simulations showing for the first time a quadruped
robot walking along a line on two point feet, and
(e) 
experimental validation of the proposed balance controller
showing the 
hydraulically actuated quadruped robot HyQ \cite{Semini2011}
balancing on a support line.

The presentation of this work is as follows: 
related work is presented in 
Sec. \ref{sec:relatedWork}; 
Section \ref{sec:balancerController} summarizes the main
concepts of the four-term balance controller which is described in
detail in \cite{Featherstone2017}.
Section \ref{sec:kinematicMapping} presents the step-by-step
procedure to obtain the kinematic mapping that extends the
controller to the quadruped case.
Section \ref{sec:locomotion} 
presents the design of the combined motion and balance
controller to achieve locomotion.
The simulations and experimental results are presented
in Sec. \ref{sec:results}, and 
the concluding remarks and topics of future work are presented in
Sec. \ref{sec:conclusion}.

\section{Related Work} \label{sec:relatedWork}
%
%


Tasks such as balancing on a line or on a single point
have been previously studied on low dimensional systems.
Some examples include
Cubli, the cube that balances and 
stands on a single point \cite{Gajamohan2012},
self-balancing bicycles \cite{Lam2011a}, 
mobile robots balancing on a ball \cite{Nagarajan2014},
and Tippy, a high-performance balancing 
robot controlled by only one actuator \cite{Sep2019}.
The balance controllers in the first two systems are 
similar in that their objective is to bring the 
systems to a completely vertical position by controlling
the rotor velocity of fly wheels.
On the other hand, the balance controller of the robot 
balancing on a ball achieves this by tracking
the desired projection of the center of mass (CoM) of the robot
using a PID controller.
The balance controller used on Tippy is based on \cite{Featherstone2017} 
and it differs
from the previously mentioned controllers
in that its goal is to track a reference angular trajectory while balancing and in
that the states of the controlled system relate to a plant describing the 
physical process of balancing, rather than directly controlling position 
and velocity states.

Generalizing the balance controllers of these systems 
to higher dimensional systems
is not a trivial task. Some of the current approaches
to achieve balancing seek to control the 
full dimensional robot by controlling  
a low dimensional virtual model while considering
the robot's centroidal 
dynamics \cite{Stephens2007a,Stephens2010a,Koolen2012}.
%
In \cite{Stephens2007a} a double inverted pendulum model 
is used to represent the actuated legs and torso
of a bipedal robot, for which combined efforts to control the
ankle and posture are synthesized. 
This effectively emulates a combined ankle 
and hip strategy for balancing. 
Other methods to achieve a balanced state consist of
taking planned steps
in locations determined by a simplified gait model, 
often also taking the form of an inverted 
pendulum \cite{Stephens2010a,Koolen2012}. 
These approaches either make use of the point mass assumption, 
hence limiting control over angular momentum, or assume
the system has access to a base of support and, thus, can 
generate ground reaction moments through the stance feet.
%
%

Other researchers have approached the problem of balancing by 
considering the full joint space dynamics.
These approaches often use the principles of
	the nominal work of \cite{Kajita2003a} on momentum-based balance control,
	which solves for desired joint motions to track specified linear 
	and angular momenta.
%
Several extensions and modifications have also been considered.
The approach in \cite{Herzog2016} instead proposes a force control
policy that considers
the coupling between the linear and angular momentum and finds appropriate
torques satisfying a hierarchy of tasks by solving
a sequence of prioritized quadratic programs (QPs).
Furthermore, the authors validate their approach 
experimentally on a torque-controlled biped.
In \cite{Henze2016} the authors opted for 
structuring the hierarchies of the tasks through 
null space projections rather than solving a series of 
QPs and solved an optimization problem only 
when determining the distribution of the balancing wrenches.

Another QP-based approach that showed a similarly challenging
balancing scenario based on partial footholds 
was presented in \cite{Wiedebach2016}. In this work, the 
authors designed a method to detect partial footholds,
i.e., foot locations in which only part of the foot is in
contact with the ground, 
and incorporated this information into their optimization
problem to design trajectories 
that successfully bring the robot to the next foothold
without falling. This allows the robot to stand on a line
contact momentarily while the swing leg 
is reaching for the next foothold 
to regain a double stance where a non-empty support polygon
allows the robot to recover its balance.
In contrast, in this work we present a
controller that not only allows the robot to balance, 
but also to linger, and 
move along a contact line, presenting a
more challenging task for the balance controller.

%
%
The aforementioned approaches 
work well on systems with a finite support polygon, but 
are impractical when balancing on a line as they
rely on controlling the center of pressure (CoP) within
the polygon,
which is now reduced to a line.
Similarly, controllers largely relying on an inaccurate CoM can  
	drive the robot to an unbalanced configuration.


For other multi-legged robots, the \emph{crawl} often constitutes the most balanced
	locomotion strategy. It is typically separated into two stages.
One stage defines a desired gait sequence and the
other plans a path for the projection of the CoM
(or, alternatively, for the zero-moment point, if this
is used as the dynamic stability criterion) such that it
traverses the sequence of support triangles formed by
the desired gait sequence 
\cite{Kim2009a, Kalakrishnan2010}. 
These approaches evidently make use of an available support polygon.
A more dynamic motion which is not based on support polygons 
is the \emph{trot}, which
involves stance phases with only two feet
in contact with the ground (thus a series of line contacts).
However,
these rely on the next stance happening soon and can still cause
the robot to fall if the gait frequency is too slow.

More recently, an approach to balance a 
	quadruped robot on two feet was proposed in \cite{Chignoli2020}. 
	The approach therein uses 
	linearization techniques on the centroidal dynamics of the robot
	to pose an optimal control problem subject to friction constraints,
	whose solution is approximated by solving a QP. 
	Our proposed approach is inherently different, 
	mainly in that we control 
	the process of balancing and the impedance associated to the 
	virtual model being used. This allows us to exploit traditional techniques to
	perform additional tasks, such
	as walking along a line while balancing.


\section{The balance controller}
\label{sec:balancerController}
\begin{figure}[b]
	\centering
	\def\svgwidth{0.95\columnwidth}
	\fontsize{8}{8}
	\executeiffilenewer{figs/balancing_physics_model.svg}{figs/balancing_physics_model.pdf}%
	{inkscape -D -z --file=figs/balancing_physics_model.svg %
	--export-pdf=figs/balancing_physics_model.pdf --export-latex}%
	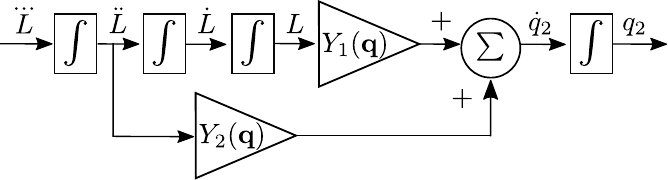%

	\caption{Plant describing the exact physics of the
		process of balancing for a 2-DoF system}
	\label{fig:balancerPlant}
\end{figure}

This section briefly summarizes the key concepts of the
balance controller presented in \cite{Featherstone2017}.
Fig. \ref{fig:balancerPlant} illustrates the plant describing
the balancing behavior of a 2-DoF planar inverted pendulum, which
swings about a pivot point on the ground (first degree of freedom)
by controlling the revolute joint connecting its two links
(second degree of freedom).
The states of the plant are the angular momentum of the
pendulum about its pivoting point, $L$,
its first and second time-derivatives, $\dot{L}$ and $\ddot{L}$,
and the position of the actuated joint used for balancing, $q_2$.
In order to achieve a balanced state,
the controller must regulate the angular momentum and its 
derivatives, and lead the joint used for balancing to its desired value.

For any planar system balancing on a point, 
there are two values
that reveal its
balancing behavior: the velocity gain 
$G_v$ and the time constant
of toppling $T_c$ \cite{Featherstone2016}. The former quantifies the
extent to which motion in the actuated joint
affects the
horizontal motion of the CoM. Hence, a small $G_v$
indicates a limited ability to control and move the CoM on top of the pivoting
point, which is required to balance.
The $T_c$ value corresponds to the time constant of the
robot falling if none of its joints moved. These two values
relate to the gains in the model of the physical process
of balancing. The particular relationship between these values
is given by
$Y_1(\mathbf{q}) = \frac{1}{mg T_c^2 G_v}$ 
and 
$Y_2(\mathbf{q}) = -\frac{1}{mg G_v}$,
where $m$ and $g$ are the total mass of the system and
the magnitude of the acceleration due to gravity, respectively,
and $\mathbf{q} \in \mathbb{R}^n$ is the vector of joint
positions of the system with $n$ degrees of freedom. Note 
that these definitions hold for a generic $n$-dimensional
planar system.

After defining the parameters of the plant describing the 
process of balancing, a proper controller can be designed. 
In \cite{Featherstone2017} such a controller is designed
by using pole placement on the plant 
assuming the parameters
$Y_1$ and $Y_2$ are constant.

It can be shown that the linear plant has a zero in the 
right-hand side of the complex plane, which implies that it 
exhibits non-minimum-phase behavior. Such behavior is intrinsic 
to the physics of balancing, and cannot be overcome by any
control system. However, it is possible to compensate
for this behavior by filtering the commanded signal
with a low-pass filter running backwards in time
(i.e., from the future to the present). This effectively
eliminates the non-minimum phase behavior from the transfer
function of the closed-loop system, resulting in motion
in which the robot \textit{leans in anticipation} of future
commanded motions. The interested reader is referred to 
\cite{Featherstone2017} for more details.

\section{Kinematic mapping}
\label{sec:kinematicMapping}
To exploit the balance controller on a quadruped robot, we define
the motion constraints that define a lower dimensional operational
space, which corresponds to the two degrees of freedom of the pendulum.
We call \emph{kinematic mapping} the function transforming the state
variables from the quadruped to the pendulum space. The mapping allows
us to apply the controller in pendulum space, and convert its output
into signals suitable for the actual robot.
The derivation of the mapping is described in the following paragraphs.

\begin{figure}[b]
	\def\svgwidth{\columnwidth}
	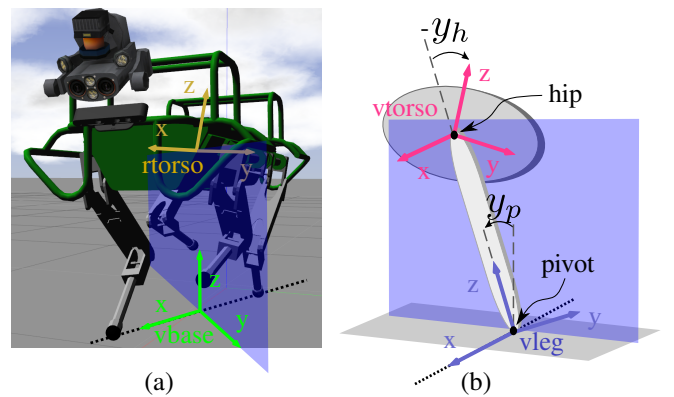
	\caption{
		Inertial and link frames of the virtual model in relation to the real robot.
		(a) The location of the stance feet dictate the 
		axis about which the joints
		of the virtual model rotate (dotted line), which is  
		orthogonal to the plane where the virtual model lies.
		(b) Location of coordinate frames of the virtual model
		and its corresponding joint variables $y_p$ and $y_h$.}
	\label{fig:hyqVirtualModel}
\end{figure}

First, consider a quadruped standing on a single pair of diagonal
legs, with the other leg pair retracted (lifted up from the ground),
as shown in Fig.~\ref{fig:hyqVirtualModel}(a). 
Project a \emph{virtual} inverted pendulum moving in the vertical
plane perpendicular to the line of contact defined by the two stance
feet. The pendulum consists of two links, the
leg and the torso, and two revolute joints, the pivot and the hip
(see Fig.~\ref{fig:hyqVirtualModel}(b)).
The pivot joint is passive and its axis coincides with the line of
contact of the quadruped; the hip joint is actuated and its axis 
is parallel to the previous one, at a distance equal
	to the length of the virtual leg link. 

The virtual torso link is composed of the real
robot torso and the retracted legs, thus they have the same inertia and
undergo the same motion (see Fig.~\ref{fig:hyqVirtualModel}(a)).
We attach three reference frames to the virtual pendulum:
\texttt{vbase} is an inertial frame with origin at the pivot
contact point with its $x$-axis coinciding with the pivot axis;
\texttt{vleg} is a frame attached to the virtual leg and also its $x$-axis
coincides with the pivot axis; \texttt{vtorso}
is attached to the virtual torso and its $x$-axis coincides with
the virtual hip axis.
The origins of \texttt{vtorso} and of the real torso frame (\texttt{rtorso})
coincide, and their $z$-axes are aligned. The frames only differ by a relative
rotation of $\phi_t$ radians about their local $z$-axis.

The states of the virtual model are
$\mathbf{y}= \left[ y_p ~ y_h \right]^T$ and
$\dot{\mathbf{y}}= \left[ \dot{y}_p ~ \dot{y}_h \right]^T$, where
$p$ and $h$ stand for pivot and hip, respectively.
The relation between the virtual states
$(\mathbf{y},\dot{\mathbf{y}})$ and the robot states
$(\mathbf{q},\dot{\mathbf{q}})$
is obtained in two steps.
First, the sensor data obtained from the robot is used to
compute the corresponding states of the virtual model 
(Sec.~\ref{subsec:sensorsToVIP}). 
Then, the kinematic mapping is derived through a 
concatenation of derived Jacobians 
(Sec.~\ref{subsec:VIPtoHYQstates}).

\subsection{Sensor data to virtual model states}
\label{subsec:sensorsToVIP}
Sensors on the robot measure the joint states $\mathbf{q}$,
$\dot{\mathbf{q}} \in \mathbb{R}^{12}$, the angular velocity of
the robot torso, $\bm{\omega} \in \mathbb{R}^3$,
and a unit vector pointing up, $\overset{\rightarrow}{\mathbf{up}}$,
both in \texttt{rtorso} coordinates. 
The stance feet position,
$\mathbf{p}_{f_i} \in \mathbb{R}^3 ~i \in \{1,2\}$,
provide the angle of the support line:
%
%
\begin{equation} \label{eq:phi_t}
\phi_t = \tan^{-1} 
	\left( \frac{p_{f_1,y} - p_{f_2,y}}{p_{f_1,x} - p_{f_2,x}} \right)
\end{equation}
where the second subscript of the foot position indicates the specific
coordinate, e.g., $p_{f_1,y}$ is the $y$-coordinate of the first stance foot.
From $\phi_t$ we can compute the rotation matrix
that transforms 3D vectors from \texttt{rtorso} to \texttt{vtorso}
coordinates, $\mathbf{E}_{\phi_t} \in SO(3)$. 
At this point, simple geometric observations yield some of the relations
between sensor data and the state of the virtual pendulum. For example,
we have that
\begin{equation} \label{eq:y1d_plus_y2d}
\dot{y}_p + \dot{y}_h = \left( \mathbf{E}_{\phi_t}  \bm{\omega} \right)_x
\end{equation}
where $(\cdot)_{x}$ indicates the $x$-component of the vector 
inside the parenthesis; we also have that
\begin{equation} \label{eq:y1_plus_y2}
y_p + y_h = \sin^{-1} 
	\left( \mathbf{E}_{\phi_t} 
			\overset{\rightarrow}{\mathbf{up}} \right)_y
\end{equation}
With reference to Fig. \ref{fig:hyqVirtualModel}(b), we also observe
that the pivot point in the \texttt{vtorso} frame, denoted by
$(p_y, p_z)$, allows us to calculate the virtual hip joint angle:
\begin{equation} \label{eq:y2_tan_fcn}
y_h = \tan^{-1} \left( \frac{p_y}{p_z} \right)
\end{equation}
and by differentiating \eqref{eq:y2_tan_fcn} w.r.t.
time, it follows that
\begin{equation} \label{eq:y2d_fcn}
\dot{y}_h = \frac{p_z  v_y - p_y  v_z}
	{p_y^2 + p_z^2}
\end{equation}
Hence, at each time instant, 
the virtual states $\mathbf{y}$ and
$\dot{\mathbf{y}}$ 
are found
by solving \eqref{eq:y1d_plus_y2d} - \eqref{eq:y2d_fcn}.
\subsection{Explicit motion constraints}
\label{subsec:VIPtoHYQstates}
The purpose of this phase is to determine the coefficients of the
following motion constraint equations:
\begin{equation} \label{eq:y_to_q}
\begin{aligned}
\dot{\mathbf{q}} &= \mathbf{G}_j \dot{\mathbf{y}} \\
\ddot{\mathbf{q}} &= \mathbf{G}_j \ddot{\mathbf{y}} + \mathbf{g}_j
\end{aligned}
\end{equation}
where $\mathbf{G}_j \in \mathbb{R}^{n \times 2}$ and
$\mathbf{g}_j = \dot{\mathbf{G}}_j \dot{\mathbf{y}}$
relate the motion of the pendulum to the exact desired motion of the real 
robot
(cf. Sec. 3.2 in \cite{Featherstone08}). 
$\mathbf{G}_j$ and $\mathbf{g}_j$ also allow us
to map the dynamics quantities of the robot,
like the inertia matrix,
into the corresponding quantity of the virtual pendulum, in order to
apply the balance controller.
Thus, the $\dot{\mathbf{q}}$ and
$\ddot{\mathbf{q}}$ in \eqref{eq:y_to_q} represent \emph{desired}
motions the robot should track in order to actually mimic the motion of
the inverted pendulum. We will therefore use these values in our control
law.

The derivation of $\mathbf{G}_j$ and $\mathbf{g}_j$ is done in stages, as
described next.

\subsubsection{Virtual pendulum to robot torso}
\label{subsubsec:VIPtoTorso}
In the first stage, we look into the relation between the virtual joint
velocity $\dot{\mathbf{y}}$ and the real torso velocity $\mathbf{v}_t$,
and we find the matrices $\mathbf{G}_t \in \mathbb{R}^{6 \times 2}$ and
$\mathbf{g}_t \in \mathbb{R}^{6}$ such that
\begin{subequations} \label{subeq:yd_to_vt}
\begin{align}
\mathbf{v}_t &= \mathbf{G}_t \dot{\mathbf{y}} \label{subeq:vt} \\
\dot{\mathbf{v}}_t &= \mathbf{G}_t \ddot{\mathbf{y}} + \mathbf{g}_t
\label{subeq:dvt}
\end{align}
\end{subequations}
Simple observation on the kinematic model of the pendulum reveals that
$\mathbf{v}_t$ can be written as follows:
\begin{equation}
\mathbf{v}_t = \mathbf{S}_p \dot{y}_p
	         + \mathbf{S}_h \dot{y}_h
\end{equation}
where $\mathbf{S}_p$ and $\mathbf{S}_h$ are the motion subspace
matrices for the pivot and hip joints, respectively.
It follows that
\begin{equation} \label{eq:Gt}
\mathbf{G}_t = \begin{bmatrix}
	\mathbf{S}_p & \mathbf{S}_h
	\end{bmatrix}
\end{equation}

To obtain $\mathbf{g}_t$, we have to compute the time
derivative of $\mathbf{G}_t$ relative to a \emph{stationary}
coordinate frame. Since the pivot axis is fixed relative to a
stationary frame, $\dot{\mathbf{S}}_p = \mathbf{0}$.
On the other hand, the hip joint axis does move in space due to the
pivot motion, thus $\mathbf{S}_h$ is changing according to
\begin{equation}
\dot{\mathbf{S}}_h = \mathbf{S}_p \dot{y}_p \times \mathbf{S}_h
\end{equation}
(cf.\@ \cite{Featherstone08}) and thus,
\begin{equation} \label{eq:gt}
\mathbf{g}_t = (\mathbf{S}_p \dot{y}_p \times \mathbf{S}_h) \dot{y}_h
\end{equation}
Note that, in the equations above, we omitted the coordinate transformation
matrices, as they do not change the algorithm but make the notation less clear.
\subsubsection{Torso to feet}
\label{subsubsec:torsoToFeet}
The goal is now to map the velocity of the torso to the velocity of the stance
feet, relative to the origin of the torso frame itself. In this phase we assume
that the stance feet are fixed on the ground.
For each stance foot $i \in \{ 1,2 \}$,
we seek the transformation matrices 
$\mathbf{G}_{f_i} \in \mathbb{R}^{3 \times 6}$ and 
$\mathbf{g}_{f_i} \in \mathbb{R}^{3}$ such that
\begin{subequations} \label{subeq:vt_to_vf}
\begin{align}
\mathbf{v}_{f_i} &= \mathbf{G}_{f_i} \mathbf{v}_t \label{subeq:vf} \\
\mathbf{a}_{f_i} &= \mathbf{G}_{f_i} \dot{\mathbf{v}}_t + \mathbf{g}_{f_i} \label{subeq:af}  
\end{align}
\end{subequations}
where $\mathbf{g}_{f_i} = \dot{\mathbf{G}}_{f_i} \mathbf{v}_t$.
Note that $\mathbf{v}_{f_i}$ and $\mathbf{a}_{f_i}$ are 3D Euclidean
vectors, as the feet of the quadruped are modeled as \emph{points}.
The velocity of the feet relative to the torso origin is the opposite
of the velocity of the torso origin relative to the foot. Thus,
\begin{equation} \label{eq:Gf}
\mathbf{G}_{f_i} = -\begin{bmatrix}
-\mathbf{p}_{f_i} \times & \mathbf{1}
\end{bmatrix}
\end{equation}
where $\mathbf{p}_{f_i} \times \in \mathbb{R}^{3 \times 3}$ is the skew-symmetric
cross product matrix of the position of the foot in torso coordinates. 
The vector $\mathbf{g}_{f_i}$ can be directly computed as
\begin{equation} \label{eq:gf}
\begin{aligned}
\mathbf{g}_{f_i} 	&= \dot{\mathbf{G}}_{f_i} \mathbf{v}_t \\
&= - \begin{bmatrix}
				-\dot{\mathbf{p}}_{f_i} \times & \mathbf{0}
				\end{bmatrix} \mathbf{v}_t \\
&= \mathbf{v}_{f_i} \times \bm{\omega}_t
\end{aligned}
\end{equation}
%
%

\subsubsection{Feet to Joint Velocities}
\label{subsubsec:feetToJointVelocities}
In the last stage we have to map feet velocity to joint velocities,
finding the matrices $\mathbf{G}_{a_i} \in \mathbb{R}^{3 \times 3}$
and $\mathbf{g}_{a_i} \in \mathbb{R}^3$ such that
\begin{subequations} \label{subeq:vf_to_qf}
\begin{align}
\dot{\mathbf{q}}_{f_i} 
	&= \mathbf{G}_{a_i} \mathbf{v}_{f_i} \label{subeq:qd_fi} \\
\ddot{\mathbf{q}}_{f_i} 
	&= \mathbf{G}_{a_i} \mathbf{a}_{f_i} + \mathbf{g}_{a_i}
\end{align}
\end{subequations}
with $\mathbf{g}_{a_i} = \dot{\mathbf{G}}_{a_i} \mathbf{v}_{f_i}$. 
Assuming that each leg of the quadruped has 3 DoFs results in
$\mathbf{G}_{a_i} = \mathbf{J}_i^{-1}$, where 
$\mathbf{J}_i$ is the Jacobian of
the stance leg $f_i$. It then follows that 
\begin{equation} \label{eq:ga}
\begin{aligned}
\mathbf{g}_{a_i} 	&= \dot{\mathbf{G}}_{a_i} \mathbf{v}_{f_i} \\
    &= \dot{\mathbf{J}}_i^{-1} \mathbf{J}_i \dot{\mathbf{q}}_{f_i}\\
    &= \left(
				-\mathbf{J}_i^{-1} \dot{\mathbf{J}}_i \mathbf{J}_i^{-1} 
				\right) \mathbf{J}_i \dot{\mathbf{q}}_{f_i} \\
  &=-\mathbf{J}_i^{-1} \dot{\mathbf{J}}_i \dot{\mathbf{q}}_{f_i}
\end{aligned}
\end{equation}
In this derivation, the formula of the derivative of the inverse
of a matrix was used from the second to the third line.

\subsubsection{Linear combination}
\label{subsubsec:linearCombination}
The transforms described in \eqref{eq:y_to_q} 
are readily constructed by collecting the
intermediate terms comprising \eqref{subeq:yd_to_vt},
\eqref{subeq:vt_to_vf}, and \eqref{subeq:vf_to_qf}.
For each stance leg $f_i$, it follows that
\begin{equation} \label{eq:qd_fi}
\dot{\mathbf{q}}_{f_i} = \mathbf{G}_{a_i} \mathbf{G}_{f_i} \mathbf{G}_t
	\dot{\mathbf{y}}
\end{equation}
and, similarly,
\begin{align}
\ddot{\mathbf{q}}_{f_i} &= \mathbf{G}_{a_i} \left( 
	\mathbf{G}_{f_i} 
		\left( 
			\mathbf{G}_{t} \ddot{\mathbf{y}} + \mathbf{g}_{t}
		\right) + \mathbf{g}_{f_i}
	\right) + \mathbf{g}_{a_i}  \nonumber \\
	&= \mathbf{G}_{a_i} \mathbf{G}_{f_i} \mathbf{G}_t \ddot{\mathbf{y}}
		+ \mathbf{G}_{a_i} \mathbf{G}_{f_i} \mathbf{g}_t 
		+ \mathbf{G}_{a_i} \mathbf{g}_{f_i}
		+ \mathbf{g}_{a_i} \label{subeq:qdd_fi}
\end{align}
%
%
%
%
%
%
By comparing \eqref{eq:qd_fi} and \eqref{subeq:qdd_fi}
with \eqref{eq:y_to_q}, it follows that
\begin{equation}\label{eq:Gj}
\mathbf{G}_j = 
\begin{bmatrix}
	\mathbf{0} \\ 
	\mathbf{G}_{a_1} \mathbf{G}_{f_1} \mathbf{G}_{t} \\
	\mathbf{G}_{a_2} \mathbf{G}_{f_2} \mathbf{G}_{t} \\
	\mathbf{0}
\end{bmatrix}
\end{equation}
and 
\begin{equation}\label{eq:gj}
\mathbf{g}_j = 
\begin{bmatrix}
	\mathbf{0} \\ 
	\mathbf{G}_{a_1} \left( \mathbf{G}_{f_1} \mathbf{g}_{t}
		+ \mathbf{g}_{f_1} \right) + \mathbf{g}_{a_1} \\
	\mathbf{G}_{a_2} \left( \mathbf{G}_{f_2} \mathbf{g}_{t} 
		+ \mathbf{g}_{f_2} \right) + \mathbf{g}_{a_2} \\
	\mathbf{0}
\end{bmatrix}
\end{equation}
for the sample case where the stance legs are the second and third leg of
the quadruped. The transforms from the pendulum states to the floating base
states are given by \eqref{subeq:yd_to_vt}.


The balance controller uses the states describing the physical process
	of balancing of the real robot and outputs $\dddot{L}$.
Using \eqref{eq:Gj} and \eqref{eq:gj} we obtain the 
numeric values of the 
equivalent 
equation of motion (EoM) of the virtual model and use these along with
	$\dddot{L}$ to compute the desired acceleration of the actuated joint, 
	$\ddot{y}_h$, to achieve this output. 
The controller on the quadruped tries to track the acceleration command
required for balancing, but also the configuration that best conforms to
the virtual pendulum model. The control law is
\begin{equation} \label{eq:tau_bal}
\bm{\tau} = 
	\mathbf{K}_P 
	\left( 
		\mathbf{q}_{vip} - \mathbf{q}
	\right)
	+ \mathbf{K}_D 
	\left( 
		\dot{\mathbf{q}}_{vip} - \dot{\mathbf{q}}
	\right)
	+ \mathbf{K}_{B} \ddot{\mathbf{q}}_{bal} 
\end{equation}
where $\ddot{\mathbf{q}}_{bal}$ is the desired joint acceleration vector,
obtained by applying \eqref{eq:y_to_q} on the pendulum acceleration
commanded by the balance controller.
The kinematic mapping \eqref{eq:y_to_q} also gives us the joint velocity
vector $\dot{\mathbf{q}}_{vip}$ corresponding to an accurate mimicking of
the pendulum behavior. The position vector $\mathbf{q}_{vip}$ is instead
constructed via inverse kinematics:
\begin{equation} \label{eq:fi_vip}
\mathbf{q}_{f_i,vip} = IK( \mathbf{p}_{f_i}(\mathbf{y}) )
\end{equation}
where $\mathbf{p}_{f_i} \in \mathbb{R}^3$ is the nominal position of the
stance foot when the quadruped is accurately mimicking the inverted pendulum.
The vector $\mathbf{q}_{vip}$ is constructed by replacing the joints of the
stance legs $f_i$ by $\mathbf{q}_{f_i, vip}$ and leaving the joints of the
swing legs unchanged.
$\mathbf{K}_P, \mathbf{K}_D, \mathbf{K}_B \in \mathbb{R}^{n \times n}$
are 
all positive definite and manually tuned based on the desired 
	error dynamics.
\section{Locomotion using the balance controller}
\label{sec:locomotion}
So far, we have presented the 
	realization of the pure task of balancing a quadruped
	by controlling a virtual 2-DoF model associated to the
	quadruped by the kinematic mapping derived in 
	Sec. \ref{sec:kinematicMapping}. Any motion foreign
	to that generated by the balance controller is 
	effectively seen as	a disturbance to the system. 
	Strictly speaking, the operational space of a robot 
	balancing while performing an additional task 
	consists of an operational
	state used for balancing, and $n-1$ operational 
	states used for the additional motion.
	
For a system with $n>2$ DoFs, the control law in \cite{Featherstone2017}
still outputs $\dddot{L}$ but is now also dependent on the motion 
of the additional states.
The EoM of the generalized
system with the additional fictitious (and static) prismatic joint is
\begin{equation} \label{eq:generalizedEOM}
\begin{bmatrix}
H_{00}	& H_{01}	& H_{02} 	& \bm{H}_{03} \\
H_{10}	& H_{11}	& H_{12} 	& \bm{H}_{13} \\
H_{20}	& H_{21}	& H_{22} 	& \bm{H}_{23} \\
\bm{H}_{30}	& \bm{H}_{31}	& \bm{H}_{32} 	& \bm{H}_{33} \\
\end{bmatrix}
\begin{bmatrix}
0 \\ \ddot{y}_p \\ \ddot{y}_h \\ \ddot{\bm{y}}_m
\end{bmatrix}
+
\begin{bmatrix}
C_0 \\ C_1 \\ C_2 \\ \bm{C}_3
\end{bmatrix}
=
\begin{bmatrix}
\tau_0 \\ 0 \\ w_h \\ \bm{w}_m
\end{bmatrix}
\end{equation}
where $\bm{y}_m$ contains the actuated generalized 
coordinates of the motion and the dots indicate
time derivatives. In fact, 
$\ddot{\bm{y}}_m$ is the output of the motion controller.
It is shown in \cite{Featherstone2017} (where all parameters
	are defined) that $\dddot{L}$ and
$\tau_0$ are directly related, hence 
\eqref{eq:generalizedEOM} can be
separated into known and unknown variables,
resulting in the system of equations
\begin{equation}\label{eq:balMotionCombined}
\begin{bmatrix}
0	& \mathbf{0} 	& H_{01}	& H_{02}  \\
0	& \mathbf{0}	& H_{11}	& H_{12}  \\
-1	& \mathbf{0}	& H_{21}	& H_{22}  \\
\mathbf{0}	& \mathbf{-1}	& \bm{H}_{31}	& \bm{H}_{32} \\
\end{bmatrix}
\begin{bmatrix}
w_h \\ \bm{w}_m \\ \ddot{y}_p \\ \ddot{y}_h
\end{bmatrix}
=
\begin{bmatrix}
-\frac{\dddot{L}}{g} - C_0 - \bm{H}_{03} \ddot{\bm{y}}_m \\ 
-C_{1} - \bm{H}_{13} \ddot{\bm{y}}_m \\ 
-C_2 - \bm{H}_{23}  \ddot{\bm{y}}_m \\ 
-\bm{C}_3 - \bm{H}_{33} \ddot{\bm{y}}_m
\end{bmatrix}
\end{equation}
which combines the outputs of the balance and the motion controllers.
After solving \eqref{eq:balMotionCombined}, 
the generalized forces $w_h$ and $\bm{w}_m$ are mapped to joint
torques through the Jacobian.

\begin{figure}
	\centering
	\def\svgwidth{0.8\columnwidth}
	\fontsize{5}{8}
	\executeiffilenewer{figs/SMFeetMotion_squareFig.svg}{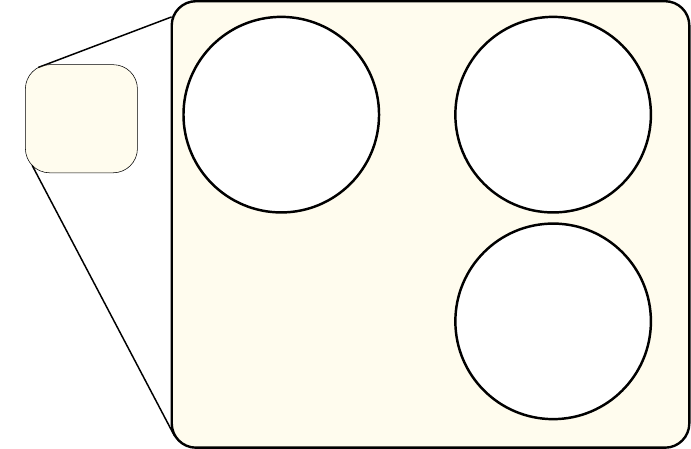}%
	{inkscape -D -z --file=figs/SMFeetMotion_squareFig.svg %
	--export-pdf=figs/SMFeetMotion_squareFig.pdf --export-latex}%
	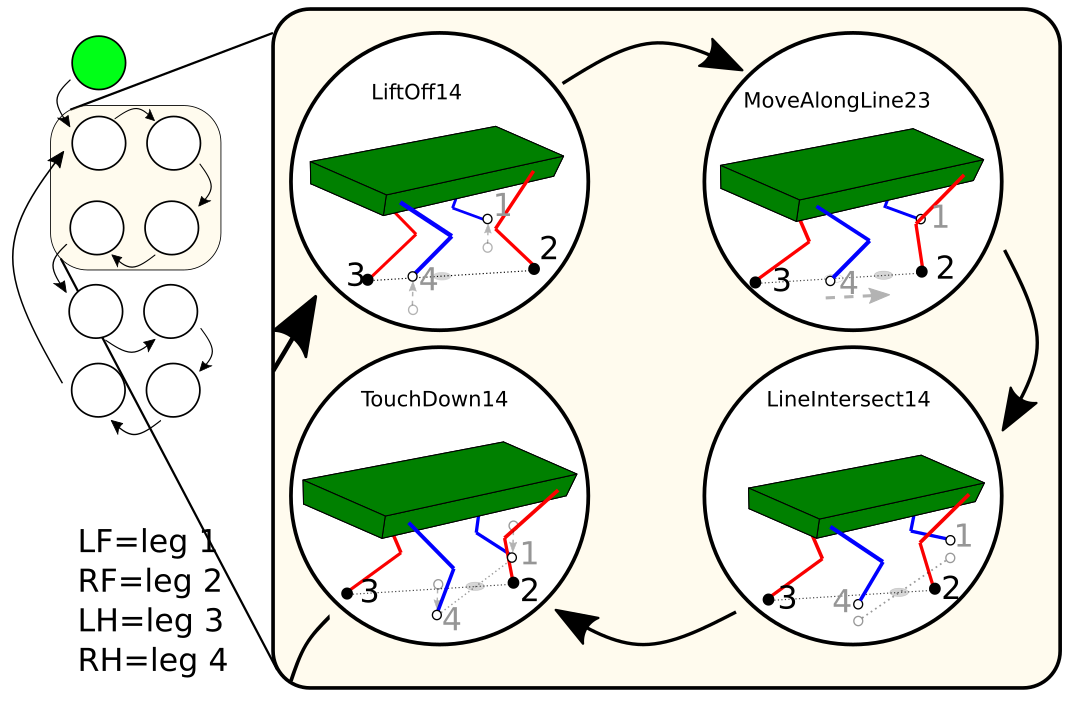%

	\caption{(Left) Modes of the state machine made to 
		generate the motion of the 
		swing feet during the balanced walk on a line.
		The $\mathrm{FullStance}$ mode (shown in bright green) 
		is executed only at the beginning. The order of the four 
		highlighted modes can be thought of as being repeated in
		the same order below them but for the LF-RH leg pair.		 
		(Right) Detail of the transitions for the RF-LH leg pair.}
	\label{fig:SMachine_feet}
\end{figure}

For the task of walking on a line while balancing, one can
	easily design motions that lie on the balance null space. 
	These are motions
	that do not disturb the CoM of the robot 
	in the direction perpendicular to the line of motion. 
	We chose to generate this motion by dividing 
	it into a series of symmetric motions executed by 
	a state machine, as shown in Fig. \ref{fig:SMachine_feet}.
The sequence of events was designed as follows.
The task is assumed to start with the quadruped having
all four legs on the ground ($\mathrm{FullStance}$). 
The state machine immediately
proceeds to the $\mathrm{LiftOff14}$ mode,
in which it lifts the LF and RH legs
and forces the balance controller
to use the RF and LH legs as its
stance legs. The robot then balances on the RF-LH leg pair and
once the robot attains a balanced state, 
it proceeds to $\mathrm{MoveAlongLine23}$. 
The user-specified distance to be traveled by the robot
is passed through a PD position controller to 
regulate the
$x$-component of the feet position, i.e., 
$\mathbf{p}_{f_i}$ in \eqref{eq:fi_vip}, effectively
bringing the robot closer to the front stance leg.
Once the robot has reached the
desired position, it moves to 
$\mathrm{LineIntersection14}$ mode, where it positions the 
swing legs 
such that the line connecting them intersects 
the current CoM $(x,y)$-position while keeping them 
slightly above the ground. Once this position has been 
reached, the $\mathrm{TouchDown14}$
event is triggered, which simply brings the swing legs down to 
touch the ground. 
The exact same procedure is then repeated for the other 
leg pair.

The balance controller was
integrated into our existing reactive control framework
(RCF) \cite{Barasuol2013a} mostly to take 
advantage of two additional tools:
\emph{gravity compensation} and the 
\emph{kinematic adjustment} to 
control the position of the feet.
The former helps remove most of the undesired terms of the 
equations of motion, leaving us with the natural task of balancing. 
The latter provides better accuracy on the positioning of the
swing legs during the touchdown event.


\section{Results}
\label{sec:results}
%
%
In this section we present the results from the tests designed
to assess the performance and robustness of the proposed strategy.
Our approach is implemented on HyQ \cite{Semini2011} and the
tests consist of different balancing tasks where the robot uses
only its diagonal leg pairs. Three tests are performed in simulation
and one experimentally (all the tests can be seen in the
accompanying video).


%
%

Our control architecture consists of a high-level and a low-level
control layer, running at 250 Hz and 1 kHz, respectively. Due to
all kinematic transformations involved, we compute the balance torques
$\mathbf{K}_{B} \mathbf{\ddot{q}}_{bal}$ in \eqref{eq:tau_bal} at the
high-level along with the $\mathbf{q}_{vip}$ and $\mathbf{\dot{q}}_{vip}$,
The latter two variables are being tracked at low-level for better impedance
rendering. The CoM states are computed using only proprioceptive sensors,
i.e. joint encoders and an Inertia Measurement Unit, that are fused in a state
estimator algorithm \cite{Nobili2017}. All of the kinematic and dynamic
transformations are generated through RobCoGen \cite{Frigerio2016}.


\subsection{Simulations} \label{sec:results_a}

Our goal with the simulations is twofold: understand the performance
of the balance controller under ideal conditions and its sensitivity to
mis-measurements and to external disturbances. For this, we designed
three simulation tests. First, we assess the controller response
when HyQ is balancing on two diagonal legs and experiences a constant
external disturbance. We chose a constant disturbance as a way of
understanding the side effect of problems like torque sensor
offset/miscalibration, inaccuracy in kinematic parameters and/or
overall robot CoM position, and also external disturbances from the hydraulic
hoses that are connected to HyQ. In a second test we evaluate the
controller performance when the robot is required to balance and
track a desired trunk trajectory at the same time. The third test
concludes the simulations with a task we called \textit{The Ninja
Walk}, where the algorithm proposed in Sec. \ref{sec:locomotion} is
evaluated in an extreme locomotion scenario.


%
%

\subsubsection{Constant disturbance} \label{sec:results_a1}
\begin{figure}
		\def\svgwidth{\columnwidth}
		\fontsize{4}{8}
	\executeiffilenewer{figs/constant_disturbance12sec.svg}{figs/constant_disturbance12sec.pdf}%
	{inkscape -D -z --file=figs/constant_disturbance12sec.svg %
	--export-pdf=figs/constant_disturbance12sec.pdf --export-latex}%
	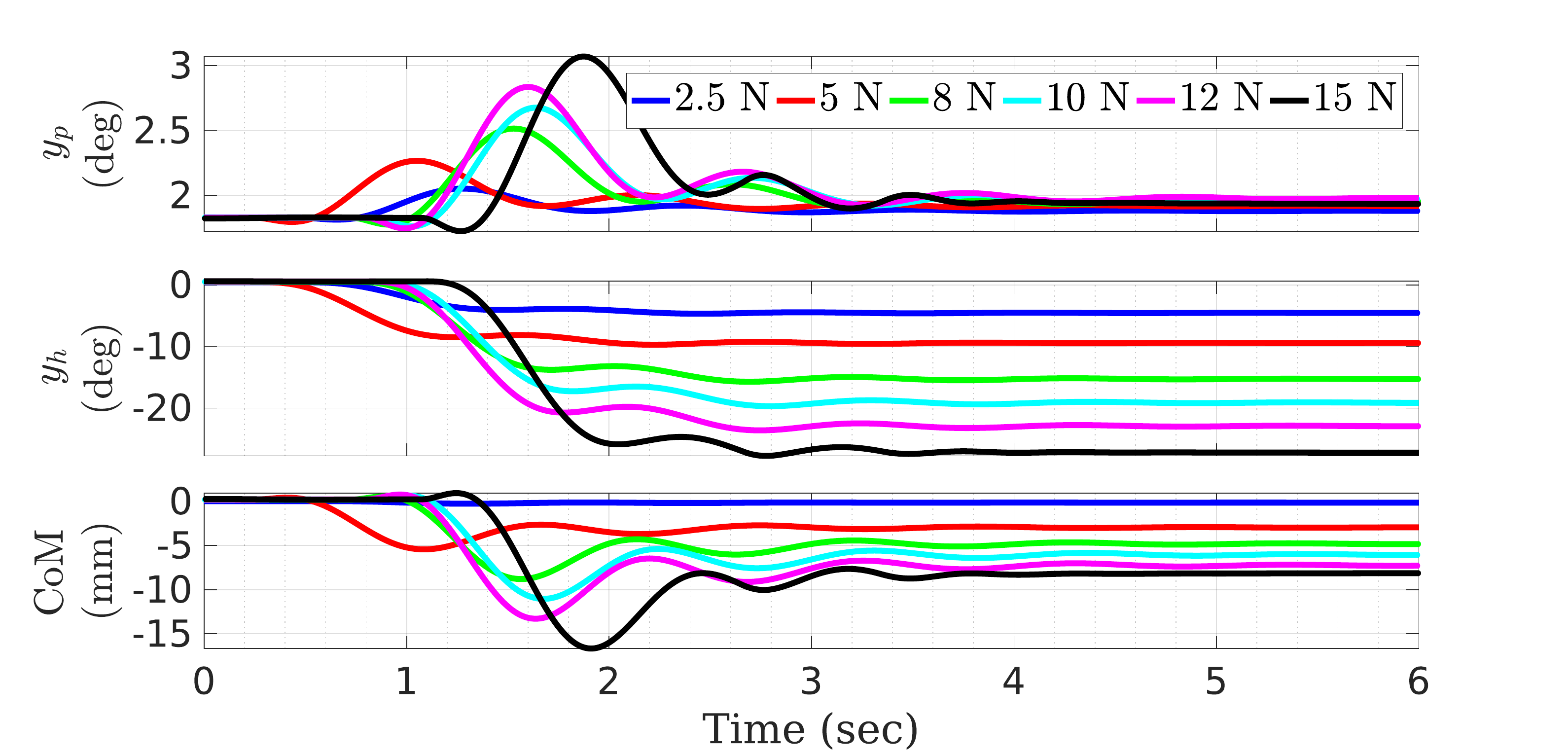%

	\caption{Trajectories of the equivalent (top) pivot joint and
		(middle) hip joint of the virtual model after HyQ 
		is pushed on the trunk with a constant force, starting
		from a balanced condition. (bottom) The corresponding
		distance of the CoM of HyQ w.r.t. the contact line,
		i.e., along the \texttt{vbase} frame. }
	\label{fig:disturbanceRejection_sim}
\end{figure}

in this test, the robot starts from a full stance condition and
proceeds to balance on two legs by simply raising a pair of diagonal
legs. When HyQ reaches a steady condition, a constant force is
applied on its floating base, in a direction perpendicular to the
line of contact formed by the stance legs (the most sensitive
orientation for the balance controller). The force is applied for
about 10 seconds until it reaches a new steady state. This process
is then repeated with increasing forces until the robot
destabilizes or touches the ground with its non-supporting legs.
The results are shown in Fig.~\ref{fig:disturbanceRejection_sim}.
It can be seen that the steady state of the virtual joint $y_h$
deviates from the desired zero value as the magnitude of the
disturbance force increases. 
This motion
	moves the CoM away from the pivot point to compensate
	for the moment created by the disturbance.
At 15 N of applied force, the robot
starts touching the ground with one of its non-supporting legs.	


%
%

\begin{figure}
	\centering
	\def\svgwidth{\columnwidth}
	\fontsize{4}{8}\fontfamily{lmss}\selectfont
	\executeiffilenewer{figs/trackingTraj_sim.svg}{figs/trackingTraj_sim.pdf}%
	{inkscape -D -z --file=figs/trackingTraj_sim.svg %
	--export-pdf=figs/trackingTraj_sim.pdf --export-latex}%
	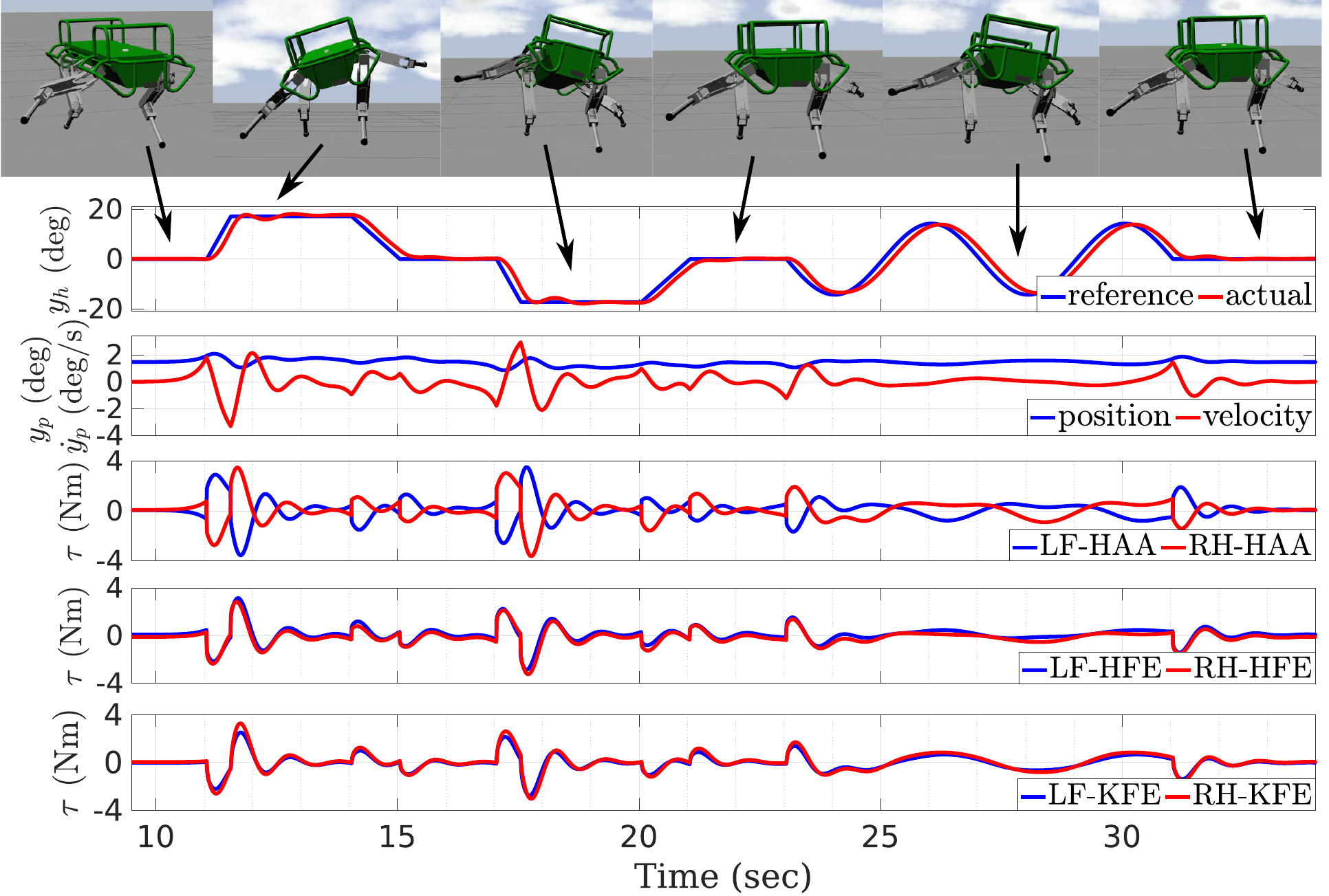%

	\fontfamily{ptm}
	\vspace*{-0.2in}
	\caption{(top) Frames of the simulation of HyQ tracking a 
		reference trajectory (second row) in which the reference
		angle is the 
		angle about the axis formed by the line connecting
		the stance legs. (third row) \textit{Leaning in anticipation}
		caused by the filtered signal. (fourth to sixth rows) Torques
		of the joints of the stance legs as computed by the control
		law \eqref{eq:tau_bal}.}
	\label{fig:trackingTask}
\end{figure}

\subsubsection{Tracking a reference trajectory} \label{sec:results_a2}

as mentioned in Sec. \ref{sec:balancerController}, 
knowledge of the reference trajectory allows us to compensate for the
inherent non-minimum phase behavior of the balancing process.  
To do so, the trajectory has to be recomputed by passing it through
a low-pass filter running backwards in time, as explained in
\cite{Featherstone2017}. The tracking task regards only the second
joint of the virtual model $y_h$. Trying to understand the balance
controller response to different input signals, we chose a reference
trajectory composed of a series of ramps with fast and slow slopes,
and a sine wave, as shown in Fig. \ref{fig:trackingTask}.
For this test, HyQ assumes a balancing configuration where the torso
(\texttt{rtorso} frame) is located 0.47 m from the ground (lower than the one used for
the disturbance test), and with the non-supporting legs wide open.
This results in a higher velocity gain that, according to the theory,
leads to a better balancing performance. The wide open posture for
the non-supporting legs is chosen to prevent them from coming into contact
with the ground during the motion of the torso.



The tracking performance and control inputs are shown in Fig.
\ref{fig:trackingTask}. The reference trajectory is tracked with a
maximum overshoot of 6\%, at the end of the steep ramps, and a milder
2\% overshoot at the end of the slower ramps. \textit{Leaning in
anticipation} can be seen by the motion of the virtual pivot
joint $y_p$ (prior to each of the ramps). This moves in the same direction of
the ramp slopes. It is important to notice that, even considering
a robot model of 90 kg, the torques required from the controller to
balance tend to be quite small. During steady states, most of the
joint efforts are carried by the gravity compensation torques.

%
%



%
%
%
%


\subsubsection{Walking across a narrow bridge (The Ninja Walk)} \label{sec:results_a3}
%
%
we have considered a narrow bridge as an extreme locomotion scenario
where footstep locations are very limited, offering very small contact
regions and creating support polygons with almost no area; in other words,
a condition where a robot would only be able to cross if balancing
on one or two legs and without depending on contact areas. The narrow
bridge is 1.5~m long and 6 cm wide, as depicted in Fig.
\ref{fig:narrow_beam}, and has the same stiffness and contact
properties as the ground in the previous simulations. For comparison,
the point-feet of HyQ are essentially spheres with a diameter of 4
cm.

\begin{figure}
	\def\svgwidth{\columnwidth}	
	\fontsize{5}{8}
	\executeiffilenewer{figs/balanced_motion.svg}{figs/balanced_motion.pdf}%
	{inkscape -D -z --file=figs/balanced_motion.svg %
	--export-pdf=figs/balanced_motion.pdf --export-latex}%
	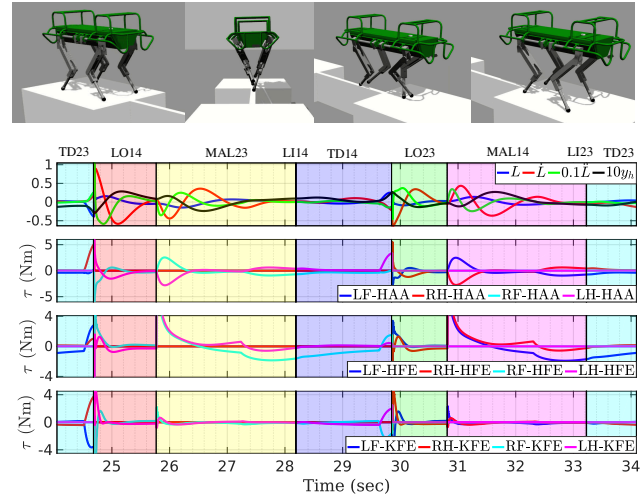%

	\vspace*{-0.2in}
	\caption{(top) Snapshots of HyQ performing the \textit{Ninja Walk} on a narrow
	bridge. (bottom) Zoomed-in plots showing the states of the plant
	converging to zero before the robot starts to move along the line.
	The shaded regions correspond to different modes of the state
	machine described in Sec. \ref{sec:locomotion}.}
	\label{fig:balancedWalk}	
	\vspace*{-0.1in}
\end{figure}

%
%
The motion results of the \textit{Ninja Walk} are detailed in Fig.
\ref{fig:balancedWalk}. HyQ successfully crosses the bridge
in about 2 minutes, counting from the initial to the final
full-stance configurations (as shown in the left and right-most
snapshots of Fig. \ref{fig:balancedWalk}). The many graphics
comprise a close-up of a full cycle of the state machine described in
Sec. \ref{sec:locomotion}, which lasts roughly 10 seconds. The
various states of the state machine are identified with different
colors, with their associated names abbreviated on the top of the
graph. For instance, the first state, abbreviated as \textit{TD23},
corresponds to the $\mathrm{TouchDown23}$ state from Fig.
\ref{fig:SMachine_feet} and is followed by $\mathrm{LiftOff14}$, then
$\mathrm{MoveAlongLine23}$, and so on. The transition events that
disturb the robot the most are $\mathrm{LO14}$, $\mathrm{LO23}$,
$\mathrm{MAL14}$ and $\mathrm{MAL23}$, as seen by the largest
variations in $\dot{L}$. Note that the lift-off events additionally
incur a slightly discontinuous jump on the position of the virtual
pendulum states as the model switches from one leg-pair to the other.
Note also that towards the end of the touch-down events, a slightly
de-stabilizing motion can be seen. This is due to the motion of the
swing legs that is generated to track a landing position in the world
frame, whose control actions are not in agreement with the balance
controller.

\subsection{Experimental validation} \label{sec:results_b}

In this section we show the preliminary results and outcomes from the
implementation of the proposed approach on a real legged machine.
The experiment consists of starting the robot in a three-leg stance
condition and then manually bringing its torso towards its balanced
position, with the balance controller activated. After HyQ balances
on its two legs, we perturb it by pushing and pulling it from its
protection frame to bring it out of balance a few times. Next, it is
left alone to regain balance. This simple experiment serves
to understand the performance
and sensitivity of the balance controller given the limitations
of the real system described in the beginning of Sec. \ref{sec:results_a}.

\begin{figure}
	\def\svgwidth{0.98\columnwidth}
	\executeiffilenewer{figs/startBalancer_exp.svg}{figs/startBalancer_exp.pdf}%
	{inkscape -D -z --file=figs/startBalancer_exp.svg %
	--export-pdf=figs/startBalancer_exp.pdf --export-latex}%
	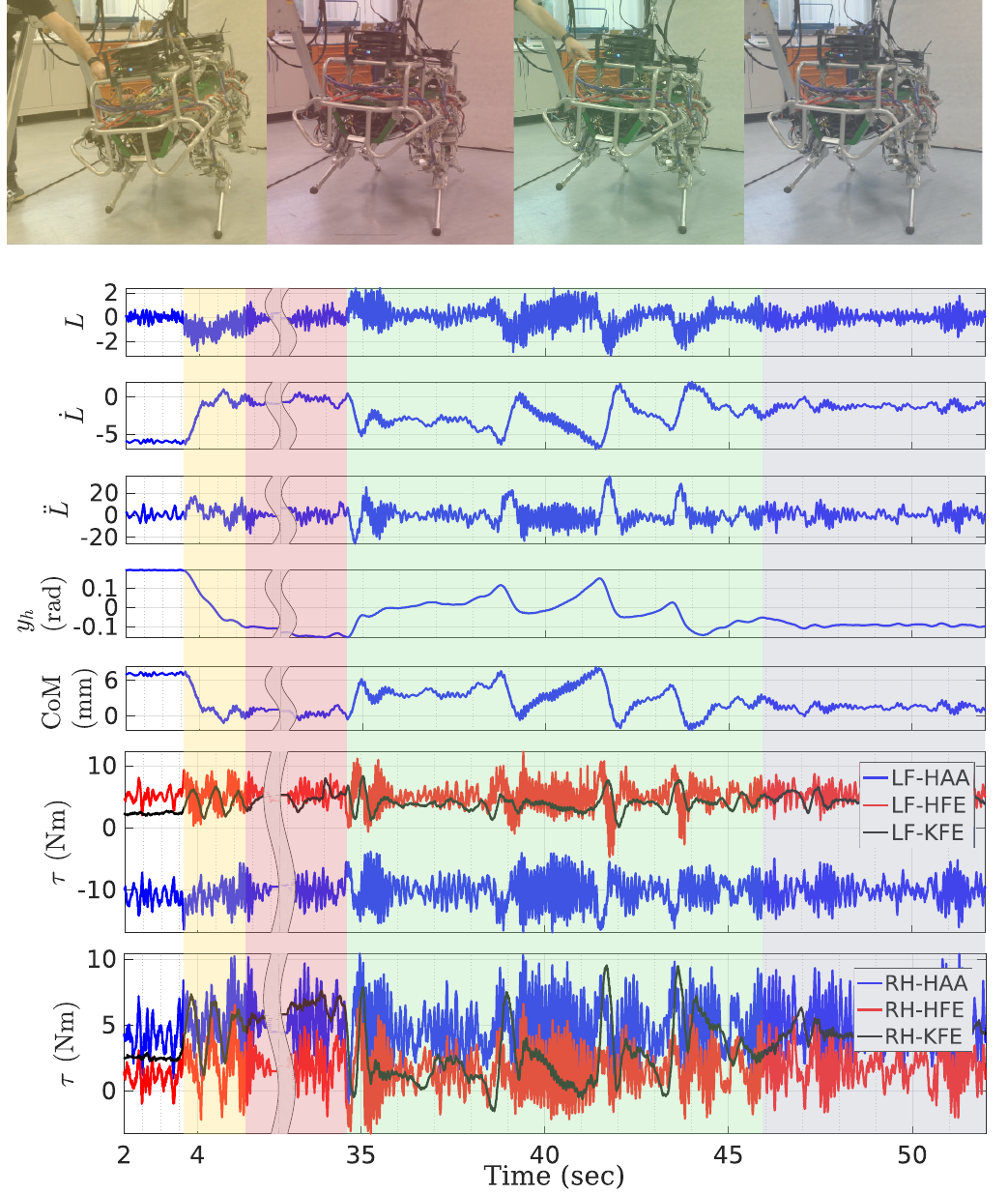%

	\caption{Snapshots and associated data of the
	balancing experimental test performed on HyQ: the robot is
	moved from three-leg to two-leg stance (yellow area); the red area
	represents balancing steady-state conditions before applying
	disturbances to the robot torso (repeated data is shrunk); series
	of impulsive disturbances are manually applied upward and downward
	to the robot torso (green area). Graphics on the bottom are states of
	the	plant in Fig. \ref{fig:balancerPlant} and the torques computed
	according to \eqref{eq:tau_bal}. (Plot made with BreakXAxis function
		created by J. Haas)} 
	\label{fig:startBalancer}
\end{figure}

The most relevant experimental data is shown in Fig.
\ref{fig:startBalancer}. The colored pictures on the top are
snapshots from the experiment that have their corresponding signals
in the bottom plots. In the translucent yellow area, it can be
seen that when the robot is moved from three-leg to two-leg stance,
the distance from the CoM to the contact line (fifth plot from top
to bottom) gets closer to zero and remains in the vicinity as it
balances alone. During a balancing steady-state, none of the
quantities converge to a fixed value. This corresponds to the natural
process of balancing, since the only way of achieving stability in an
inherently unstable system is keeping it in motion. Note, however, that the
quantities $L$, $\dot{L}$ and the CoM distance to the support line do
not oscillate around zero. This behavior resembles the steady-state
conditions observed in the constant disturbance test, shown in Sec.
\ref{sec:results_a1}, where the disturbance in this case are mainly
related to: the inaccuracy of torque sensors; a wrong
estimation of the CoM position; and the external forces and moments
from the hydraulic hoses connected to HyQ. The latter was clearly
observed during the experiments. The mechanical flexibility of the
robot structure is part of the non-modeled dynamics that constitutes
a relevant source of disturbances that substantially affects the
estimation of the CoM related states.

The translucent green area in Fig.~\ref{fig:startBalancer}
corresponds to the manually applied impulsive disturbances, which can
be seen from the big variations in $y_h$, which is closely related to
the motion of the torso. After the impulsive disturbances, the robot
comes back to another balancing steady-state (translucent blue area). 

\section{Conclusions}
\label{sec:conclusion}

This work proposes a balancing approach that does not rely on support
polygons and can be applied to legged robots that have flat-feet or
even point-feet. The approach is based on the balancing method
presented in \cite{Featherstone2017} used to balance a 2-DoF
virtual model, and a kinematic mapping to transform the virtual
model quantities into the robot joint space. Furthermore, we showed
that a rather simple motion generation algorithm combined with the
proposed balancing approach would allow a robot to walk over
extremely difficult scenarios. Simulation results were designed
to assess the controller performance
and to understand its sensitivity to disturbances that can be
encountered in a real implementation. We presented preliminary
experimental results showing for the first time a point-feet
quadruped robot balancing on two legs. Both simulation and
experimental tests were very useful to bring knowledge on the
requirements to achieve good balancing performance. The most
relevant findings are: a) it is important having accurate torque
measurements (for torque-controlled robots), since the balance
controller generates low torque commands; the reference tracking
task is very sensitive to disturbances, that can be represented by
torque offsets, external forces or biased CoM estimation; and any
non-modelled flexibility along the robot kinematic structure can
introduce unrealistic CoM measurements.

Future work considers improving the balance performance by extending
the virtual model to also output commands to the non-supporting legs.

\bibliographystyle{style/IEEEtran}
\bibliography{myPaperRefs-ral2020}

\end{document}

%% file: figs/balancing_physics_model.pdf_tex
\begingroup%
  \makeatletter%
  \providecommand\color[2][]{%
    \errmessage{(Inkscape) Color is used for the text in Inkscape, but the package 'color.sty' is not loaded}%
    \renewcommand\color[2][]{}%
  }%
  \providecommand\transparent[1]{%
    \errmessage{(Inkscape) Transparency is used (non-zero) for the text in Inkscape, but the package 'transparent.sty' is not loaded}%
    \renewcommand\transparent[1]{}%
  }%
  \providecommand\rotatebox[2]{#2}%
  \newcommand*\fsize{\dimexpr\f@size pt\relax}%
  \newcommand*\lineheight[1]{\fontsize{\fsize}{#1\fsize}\selectfont}%
  \ifx\svgwidth\undefined%
    \setlength{\unitlength}{192.15736595bp}%
    \ifx\svgscale\undefined%
      \relax%
    \else%
      \setlength{\unitlength}{\unitlength * \real{\svgscale}}%
    \fi%
  \else%
    \setlength{\unitlength}{\svgwidth}%
  \fi%
  \global\let\svgwidth\undefined%
  \global\let\svgscale\undefined%
  \makeatother%
  \begin{picture}(1,0.2691828)%
    \lineheight{1}%
    \setlength\tabcolsep{0pt}%
    \put(0,0){\includegraphics[width=\unitlength,page=1]{balancing_physics_model.pdf}}%
  \end{picture}%
\endgroup%

%% file: figs/hyq_vip_frames3D.pdf_tex
\begingroup%
  \makeatletter%
  \providecommand\color[2][]{%
    \errmessage{(Inkscape) Color is used for the text in Inkscape, but the package 'color.sty' is not loaded}%
    \renewcommand\color[2][]{}%
  }%
  \providecommand\transparent[1]{%
    \errmessage{(Inkscape) Transparency is used (non-zero) for the text in Inkscape, but the package 'transparent.sty' is not loaded}%
    \renewcommand\transparent[1]{}%
  }%
  \providecommand\rotatebox[2]{#2}%
  \newcommand*\fsize{\dimexpr\f@size pt\relax}%
  \newcommand*\lineheight[1]{\fontsize{\fsize}{#1\fsize}\selectfont}%
  \ifx\svgwidth\undefined%
    \setlength{\unitlength}{271.56046316bp}%
    \ifx\svgscale\undefined%
      \relax%
    \else%
      \setlength{\unitlength}{\unitlength * \real{\svgscale}}%
    \fi%
  \else%
    \setlength{\unitlength}{\svgwidth}%
  \fi%
  \global\let\svgwidth\undefined%
  \global\let\svgscale\undefined%
  \makeatother%
  \begin{picture}(1,0.59292599)%
    \lineheight{1}%
    \setlength\tabcolsep{0pt}%
    \put(0,0){\includegraphics[width=\unitlength,page=1]{figs/hyq_vip_frames3D.pdf}}%
    \put(0.35255975,0.33508655){\color[rgb]{0.78431373,0.67058824,0.21568627}\rotatebox{0.52770676}{\makebox(0,0)[lt]{\lineheight{0}\smash{\begin{tabular}[t]{l}y\end{tabular}}}}}%
    \put(0,0){\includegraphics[width=\unitlength,page=2]{figs/hyq_vip_frames3D.pdf}}%
    \put(0.20537571,0.00582571){\color[rgb]{0,0,0}\makebox(0,0)[lt]{\lineheight{1.25}\smash{\begin{tabular}[t]{l}(a)\end{tabular}}}}%
    \put(0.6921452,0.00582571){\color[rgb]{0,0,0}\makebox(0,0)[lt]{\lineheight{1.25}\smash{\begin{tabular}[t]{l}(b)\end{tabular}}}}%
    \put(0.21985826,0.07752078){\color[rgb]{0,1,0}\makebox(0,0)[lt]{\lineheight{0}\smash{\begin{tabular}[t]{l}vbase\end{tabular}}}}%
    \put(0.2015236,0.33896412){\color[rgb]{0.83137255,0.66666667,0}\makebox(0,0)[lt]{\lineheight{0}\smash{\begin{tabular}[t]{l}rtorso\end{tabular}}}}%
    \put(0.30469327,0.16384951){\color[rgb]{0,1,0}\rotatebox{0.52770676}{\makebox(0,0)[lt]{\lineheight{0}\smash{\begin{tabular}[t]{l}z\end{tabular}}}}}%
    \put(0,0){\includegraphics[width=\unitlength,page=3]{figs/hyq_vip_frames3D.pdf}}%
    \put(0.22173578,0.12521754){\color[rgb]{0,1,0}\rotatebox{0.52770676}{\makebox(0,0)[lt]{\lineheight{0}\smash{\begin{tabular}[t]{l}x\end{tabular}}}}}%
    \put(0.34580729,0.10240358){\color[rgb]{0,1,0}\rotatebox{0.52770676}{\makebox(0,0)[lt]{\lineheight{0}\smash{\begin{tabular}[t]{l}y\end{tabular}}}}}%
    \put(0.2655071,0.46189392){\color[rgb]{0.78431373,0.67058824,0.21568627}\rotatebox{0.52770676}{\makebox(0,0)[lt]{\lineheight{0}\smash{\begin{tabular}[t]{l}z\end{tabular}}}}}%
    \put(0.22247981,0.39613134){\color[rgb]{0.78431373,0.67058824,0.21568627}\rotatebox{0.52770676}{\makebox(0,0)[lt]{\lineheight{0}\smash{\begin{tabular}[t]{l}x\end{tabular}}}}}%
    \put(0,0){\includegraphics[width=\unitlength,page=4]{figs/hyq_vip_frames3D.pdf}}%
    \put(0.66542057,0.06948421){\color[rgb]{0.4,0.4,0.8}\rotatebox{0.52770619}{\makebox(0,0)[lt]{\lineheight{0}\smash{\begin{tabular}[t]{l}x\end{tabular}}}}}%
    \put(0.78195019,0.06592018){\color[rgb]{0.24705882,0.24705882,0.74509804}\makebox(0,0)[lt]{\lineheight{0}\smash{\begin{tabular}[t]{l}vleg\end{tabular}}}}%
    \put(0.73092357,0.34053557){\color[rgb]{1,0.21568627,0.56078431}\rotatebox{-4.9460144}{\makebox(0,0)[lt]{\lineheight{0}\smash{\begin{tabular}[t]{l}y\end{tabular}}}}}%
    \put(0.62583014,0.33085152){\color[rgb]{1,0.21568627,0.56078431}\rotatebox{0.64816725}{\makebox(0,0)[lt]{\lineheight{0}\smash{\begin{tabular}[t]{l}x\end{tabular}}}}}%
    \put(0.7202658,0.4807188){\color[rgb]{1,0.21568627,0.56078431}\rotatebox{-0.38579945}{\makebox(0,0)[lt]{\lineheight{0}\smash{\begin{tabular}[t]{l}z\end{tabular}}}}}%
    \put(0.55559671,0.42685317){\color[rgb]{1,0.21568627,0.56078431}\makebox(0,0)[lt]{\lineheight{0}\smash{\begin{tabular}[t]{l}vtorso\end{tabular}}}}%
    \put(0,0){\includegraphics[width=\unitlength,page=5]{figs/hyq_vip_frames3D.pdf}}%
    \put(0.63,0.5505681){\color[rgb]{0,0,0}\makebox(0,0)[lt]{\lineheight{1.25}\smash{\begin{tabular}[t]{l}-\end{tabular}}}}%
    \put(0.70197043,0.15575557){\color[rgb]{0.4,0.4,0.8}\rotatebox{-0.50625834}{\makebox(0,0)[lt]{\lineheight{0}\smash{\begin{tabular}[t]{l}z\end{tabular}}}}}%
    \put(0.88780274,0.10666482){\color[rgb]{0.4,0.4,0.8}\rotatebox{-5.0664731}{\makebox(0,0)[lt]{\lineheight{0}\smash{\begin{tabular}[t]{l}y\end{tabular}}}}}%
    \put(0,0){\includegraphics[width=\unitlength,page=6]{figs/hyq_vip_frames3D.pdf}}%
    \put(0.81635651,0.18639581){\color[rgb]{0,0,0}\makebox(0,0)[lt]{\lineheight{1.25}\smash{\begin{tabular}[t]{l}pivot\end{tabular}}}}%
    \put(0.82765465,0.44826299){\color[rgb]{0,0,0}\makebox(0,0)[lt]{\lineheight{1.25}\smash{\begin{tabular}[t]{l}hip\end{tabular}}}}%
    \put(0,0){\includegraphics[width=\unitlength,page=7]{figs/hyq_vip_frames3D.pdf}}%
  \end{picture}%
\endgroup%

%% file: figs/SMFeetMotion_squareFig.pdf_tex
\begingroup%
  \makeatletter%
  \providecommand\color[2][]{%
    \errmessage{(Inkscape) Color is used for the text in Inkscape, but the package 'color.sty' is not loaded}%
    \renewcommand\color[2][]{}%
  }%
  \providecommand\transparent[1]{%
    \errmessage{(Inkscape) Transparency is used (non-zero) for the text in Inkscape, but the package 'transparent.sty' is not loaded}%
    \renewcommand\transparent[1]{}%
  }%
  \providecommand\rotatebox[2]{#2}%
  \newcommand*\fsize{\dimexpr\f@size pt\relax}%
  \newcommand*\lineheight[1]{\fontsize{\fsize}{#1\fsize}\selectfont}%
  \ifx\svgwidth\undefined%
    \setlength{\unitlength}{198.873495bp}%
    \ifx\svgscale\undefined%
      \relax%
    \else%
      \setlength{\unitlength}{\unitlength * \real{\svgscale}}%
    \fi%
  \else%
    \setlength{\unitlength}{\svgwidth}%
  \fi%
  \global\let\svgwidth\undefined%
  \global\let\svgscale\undefined%
  \makeatother%
  \begin{picture}(1,0.65004678)%
    \lineheight{1}%
    \setlength\tabcolsep{0pt}%
    \put(0,0){\includegraphics[width=\unitlength,page=1]{SMFeetMotion_squareFig.pdf}}%
    \put(0.71845894,0.26988194){\color[rgb]{0,0,0}\makebox(0,0)[lt]{\lineheight{1.25}\smash{\begin{tabular}[t]{l}LineIntersect14\end{tabular}}}}%
    \put(0,0){\includegraphics[width=\unitlength,page=2]{SMFeetMotion_squareFig.pdf}}%
    \put(0.34203463,0.56234264){\color[rgb]{0,0,0}\makebox(0,0)[lt]{\lineheight{1.25}\smash{\begin{tabular}[t]{l}LiftOff14\end{tabular}}}}%
    \put(0.69649917,0.55449077){\color[rgb]{0,0,0}\makebox(0,0)[lt]{\lineheight{1.25}\smash{\begin{tabular}[t]{l}MoveAlongLine23\end{tabular}}}}%
    \put(0.3324437,0.26996144){\color[rgb]{0,0,0}\makebox(0,0)[lt]{\lineheight{1.25}\smash{\begin{tabular}[t]{l}TouchDown14\end{tabular}}}}%
    \put(0,0){\includegraphics[width=\unitlength,page=3]{SMFeetMotion_squareFig.pdf}}%
    \put(0.4579207,0.45121806){\color[rgb]{0.58823529,0.58823529,0.58823529}\makebox(0,0)[lt]{\lineheight{1.25}\smash{\begin{tabular}[t]{l}1\end{tabular}}}}%
    \put(0.50204811,0.4100824){\color[rgb]{0,0,0}\makebox(0,0)[lt]{\lineheight{1.25}\smash{\begin{tabular}[t]{l}2\end{tabular}}}}%
    \put(0.31793893,0.38477098){\color[rgb]{0,0,0}\makebox(0,0)[lt]{\lineheight{1.25}\smash{\begin{tabular}[t]{l}3\end{tabular}}}}%
    \put(0.38736793,0.38046226){\color[rgb]{0.58823529,0.58823529,0.58823529}\makebox(0,0)[lt]{\lineheight{1.25}\smash{\begin{tabular}[t]{l}4\end{tabular}}}}%
    \put(0.87455943,0.44004444){\color[rgb]{0.58823529,0.58823529,0.58823529}\makebox(0,0)[lt]{\lineheight{1.25}\smash{\begin{tabular}[t]{l}1\end{tabular}}}}%
    \put(0.87962752,0.39150103){\color[rgb]{0,0,0}\makebox(0,0)[lt]{\lineheight{1.25}\smash{\begin{tabular}[t]{l}2\end{tabular}}}}%
    \put(0.72366797,0.37965808){\color[rgb]{0,0,0}\makebox(0,0)[lt]{\lineheight{1.25}\smash{\begin{tabular}[t]{l}3\end{tabular}}}}%
    \put(0.78750744,0.37689823){\color[rgb]{0.58823529,0.58823529,0.58823529}\makebox(0,0)[lt]{\lineheight{1.25}\smash{\begin{tabular}[t]{l}4\end{tabular}}}}%
    \put(0.48329318,0.11948887){\color[rgb]{0.58823529,0.58823529,0.58823529}\makebox(0,0)[lt]{\lineheight{1.25}\smash{\begin{tabular}[t]{l}1\end{tabular}}}}%
    \put(0.48364726,0.08441426){\color[rgb]{0,0,0}\makebox(0,0)[lt]{\lineheight{1.25}\smash{\begin{tabular}[t]{l}2\end{tabular}}}}%
    \put(0.3310548,0.08334638){\color[rgb]{0,0,0}\makebox(0,0)[lt]{\lineheight{1.25}\smash{\begin{tabular}[t]{l}3\end{tabular}}}}%
    \put(0.40937316,0.05667955){\color[rgb]{0.58823529,0.58823529,0.58823529}\makebox(0,0)[lt]{\lineheight{1.25}\smash{\begin{tabular}[t]{l}4\end{tabular}}}}%
    \put(0.8967828,0.13632454){\color[rgb]{0.58823529,0.58823529,0.58823529}\makebox(0,0)[lt]{\lineheight{1.25}\smash{\begin{tabular}[t]{l}1\end{tabular}}}}%
    \put(0.885015,0.08643453){\color[rgb]{0,0,0}\makebox(0,0)[lt]{\lineheight{1.25}\smash{\begin{tabular}[t]{l}2\end{tabular}}}}%
    \put(0.72905545,0.07728533){\color[rgb]{0,0,0}\makebox(0,0)[lt]{\lineheight{1.25}\smash{\begin{tabular}[t]{l}3\end{tabular}}}}%
    \put(0.78110993,0.07385179){\color[rgb]{0.58823529,0.58823529,0.58823529}\makebox(0,0)[lt]{\lineheight{1.25}\smash{\begin{tabular}[t]{l}4\end{tabular}}}}%
    \put(0,0){\includegraphics[width=\unitlength,page=4]{SMFeetMotion_squareFig.pdf}}%
    \put(0.06242767,0.13106952){\color[rgb]{0,0,0}\makebox(0,0)[lt]{\lineheight{1.25}\smash{\begin{tabular}[t]{l}LF=leg 1\\RF=leg 2\\LH=leg 3\\RH=leg 4\end{tabular}}}}%
    \put(1.27649906,-4.9534835){\color[rgb]{0,0,0}\makebox(0,0)[lt]{\begin{minipage}{0.61850064\unitlength}\raggedright \end{minipage}}}%
  \end{picture}%
\endgroup%

%% file: figs/constant_disturbance12sec.pdf_tex
\begingroup%
  \makeatletter%
  \providecommand\color[2][]{%
    \errmessage{(Inkscape) Color is used for the text in Inkscape, but the package 'color.sty' is not loaded}%
    \renewcommand\color[2][]{}%
  }%
  \providecommand\transparent[1]{%
    \errmessage{(Inkscape) Transparency is used (non-zero) for the text in Inkscape, but the package 'transparent.sty' is not loaded}%
    \renewcommand\transparent[1]{}%
  }%
  \providecommand\rotatebox[2]{#2}%
  \newcommand*\fsize{\dimexpr\f@size pt\relax}%
  \newcommand*\lineheight[1]{\fontsize{\fsize}{#1\fsize}\selectfont}%
  \ifx\svgwidth\undefined%
    \setlength{\unitlength}{915.75bp}%
    \ifx\svgscale\undefined%
      \relax%
    \else%
      \setlength{\unitlength}{\unitlength * \real{\svgscale}}%
    \fi%
  \else%
    \setlength{\unitlength}{\svgwidth}%
  \fi%
  \global\let\svgwidth\undefined%
  \global\let\svgscale\undefined%
  \makeatother%
  \begin{picture}(1,0.48075348)%
    \lineheight{1}%
    \setlength\tabcolsep{0pt}%
    \put(0,0){\includegraphics[width=\unitlength,page=1]{figs/constant_disturbance12sec.pdf}}%
  \end{picture}%
\endgroup%

%% file: figs/trackingTraj_sim.pdf_tex
\begingroup%
  \makeatletter%
  \providecommand\color[2][]{%
    \errmessage{(Inkscape) Color is used for the text in Inkscape, but the package 'color.sty' is not loaded}%
    \renewcommand\color[2][]{}%
  }%
  \providecommand\transparent[1]{%
    \errmessage{(Inkscape) Transparency is used (non-zero) for the text in Inkscape, but the package 'transparent.sty' is not loaded}%
    \renewcommand\transparent[1]{}%
  }%
  \providecommand\rotatebox[2]{#2}%
  \newcommand*\fsize{\dimexpr\f@size pt\relax}%
  \newcommand*\lineheight[1]{\fontsize{\fsize}{#1\fsize}\selectfont}%
  \ifx\svgwidth\undefined%
    \setlength{\unitlength}{556.35400004bp}%
    \ifx\svgscale\undefined%
      \relax%
    \else%
      \setlength{\unitlength}{\unitlength * \real{\svgscale}}%
    \fi%
  \else%
    \setlength{\unitlength}{\svgwidth}%
  \fi%
  \global\let\svgwidth\undefined%
  \global\let\svgscale\undefined%
  \makeatother%
  \begin{picture}(1,0.67401209)%
    \lineheight{1}%
    \setlength\tabcolsep{0pt}%
    \put(0,0){\includegraphics[width=\unitlength,page=1]{figs/trackingTraj_sim.pdf}}%
  \end{picture}%
\endgroup%

%% file: figs/balanced_motion.pdf_tex
\begingroup%
  \makeatletter%
  \providecommand\color[2][]{%
    \errmessage{(Inkscape) Color is used for the text in Inkscape, but the package 'color.sty' is not loaded}%
    \renewcommand\color[2][]{}%
  }%
  \providecommand\transparent[1]{%
    \errmessage{(Inkscape) Transparency is used (non-zero) for the text in Inkscape, but the package 'transparent.sty' is not loaded}%
    \renewcommand\transparent[1]{}%
  }%
  \providecommand\rotatebox[2]{#2}%
  \newcommand*\fsize{\dimexpr\f@size pt\relax}%
  \newcommand*\lineheight[1]{\fontsize{\fsize}{#1\fsize}\selectfont}%
  \ifx\svgwidth\undefined%
    \setlength{\unitlength}{511.24709575bp}%
    \ifx\svgscale\undefined%
      \relax%
    \else%
      \setlength{\unitlength}{\unitlength * \real{\svgscale}}%
    \fi%
  \else%
    \setlength{\unitlength}{\svgwidth}%
  \fi%
  \global\let\svgwidth\undefined%
  \global\let\svgscale\undefined%
  \makeatother%
  \begin{picture}(1,0.75756391)%
    \lineheight{1}%
    \setlength\tabcolsep{0pt}%
    \put(0,0){\includegraphics[width=\unitlength,page=1]{figs/balanced_motion.pdf}}%
    \put(0.07325012,0.52160692){\color[rgb]{0,0,0}\makebox(0,0)[lt]{\lineheight{1.25}\smash{\begin{tabular}[t]{l}TD23\end{tabular}}}}%
    \put(0.15407492,0.52001612){\color[rgb]{0,0,0}\makebox(0,0)[lt]{\lineheight{1.25}\smash{\begin{tabular}[t]{l}LO14\end{tabular}}}}%
    \put(0.2995048,0.52021342){\color[rgb]{0,0,0}\makebox(0,0)[lt]{\lineheight{1.25}\smash{\begin{tabular}[t]{l}MAL23\end{tabular}}}}%
    \put(0.41839433,0.52021342){\color[rgb]{0,0,0}\makebox(0,0)[lt]{\lineheight{1.25}\smash{\begin{tabular}[t]{l}LI14\end{tabular}}}}%
    \put(0.48642805,0.51903011){\color[rgb]{0,0,0}\makebox(0,0)[lt]{\lineheight{1.25}\smash{\begin{tabular}[t]{l}TD14\end{tabular}}}}%
    \put(0.60596293,0.52021342){\color[rgb]{0,0,0}\makebox(0,0)[lt]{\lineheight{1.25}\smash{\begin{tabular}[t]{l}LO23\end{tabular}}}}%
    \put(0.73224828,0.52119473){\color[rgb]{0,0,0}\makebox(0,0)[lt]{\lineheight{1.25}\smash{\begin{tabular}[t]{l}MAL14\end{tabular}}}}%
    \put(0.85665095,0.52098762){\color[rgb]{0,0,0}\makebox(0,0)[lt]{\lineheight{1.25}\smash{\begin{tabular}[t]{l}LI23\end{tabular}}}}%
    \put(0.91206924,0.52020386){\color[rgb]{0,0,0}\makebox(0,0)[lt]{\lineheight{1.25}\smash{\begin{tabular}[t]{l}TD23\end{tabular}}}}%
  \end{picture}%
\endgroup%

%% file: figs/startBalancer_exp.pdf_tex
\begingroup%
  \makeatletter%
  \providecommand\color[2][]{%
    \errmessage{(Inkscape) Color is used for the text in Inkscape, but the package 'color.sty' is not loaded}%
    \renewcommand\color[2][]{}%
  }%
  \providecommand\transparent[1]{%
    \errmessage{(Inkscape) Transparency is used (non-zero) for the text in Inkscape, but the package 'transparent.sty' is not loaded}%
    \renewcommand\transparent[1]{}%
  }%
  \providecommand\rotatebox[2]{#2}%
  \newcommand*\fsize{\dimexpr\f@size pt\relax}%
  \newcommand*\lineheight[1]{\fontsize{\fsize}{#1\fsize}\selectfont}%
  \ifx\svgwidth\undefined%
    \setlength{\unitlength}{300.30028095bp}%
    \ifx\svgscale\undefined%
      \relax%
    \else%
      \setlength{\unitlength}{\unitlength * \real{\svgscale}}%
    \fi%
  \else%
    \setlength{\unitlength}{\svgwidth}%
  \fi%
  \global\let\svgwidth\undefined%
  \global\let\svgscale\undefined%
  \makeatother%
  \begin{picture}(1,1.18356489)%
    \lineheight{1}%
    \setlength\tabcolsep{0pt}%
    \put(0,0){\includegraphics[width=\unitlength,page=1]{figs/startBalancer_exp.pdf}}%
  \end{picture}%
\endgroup%